\documentclass{article}

\usepackage{arxiv}

\usepackage[utf8]{inputenc} 
\usepackage[T1]{fontenc}    
\usepackage{hyperref}       
\usepackage{url}            
\usepackage{booktabs}       
\usepackage{amsfonts}       
\usepackage{nicefrac}       
\usepackage{microtype}      
\usepackage{lipsum}
\usepackage{graphicx}
\usepackage{subcaption} 
\usepackage[numbers]{natbib}
\usepackage{amsmath}
\usepackage{algorithm}
\usepackage{lipsum}
\usepackage{algorithmicx}
\usepackage{algpseudocode}
\graphicspath{ {./images/} }

\title{Policy Complexity, Reaction Time, and Bounded Rationality in Reinforcement Learning}

\author{
 James Wu \\
  Department of Computer Science\\
  Rensselaer Polytechnic Institute\\
  Troy, NY, 12180 \\
  \texttt{wuj27@rpi.edu} \\
   \And
 Chris R. Sims \\
  Department of Cognitive Science\\
  Rensselaer Polytechnic Institute\\
  Troy, NY, 12180 \\
  \texttt{simsc3@rpi.edu} \\
}

\begin{document}
\maketitle
\begin{abstract}
Biological agents do not learn under conditions of unlimited computation. For humans, learning and choice are shaped by constraints on perception, attention, and working memory, which limit how much state information guides behavior and therefore bound policy complexity. Standard reinforcement learning models typically optimize reward without explicitly representing these internal costs, making them less suitable as models of biological intelligence. We derive MI-SARSA, an on-policy temporal-difference algorithm that incorporates mutual-information regularization through a learned marginal action prior and a penalty on state-specific deviations from that prior. This yields a sequential learning model in which state information is used selectively when its expected return benefit justifies the added informational cost. Critically, the same state-specific information cost that governs policy compression also generates trial-level predictions for reaction time, distinguishing MI-SARSA from most reinforcement learning models, which predict choices or returns but not latency. Empirically, MI-SARSA produces a reward-complexity tradeoff, and stronger information penalties produce simpler policies with lower control costs and faster reaction times. Under environment shift, increasing regularization reduces post-switch performance degradation but also lowers asymptotic return, revealing a robustness-capacity tradeoff. Together, these results position MI-SARSA as a model of bounded sequential learning under cognitive constraints.
\end{abstract}


\section{Introduction}
Biological learning and choice are shaped by severe constraints on perception, attention, and working memory \citep{miller1956,broadbent1958}. These constraints limit how much information an organism can use to condition behavior on the current state, and therefore place a bound on the complexity of the policy that can be implemented in practice. Standard reinforcement learning models, however, typically treat decision policies as unconstrained mappings from states to actions, optimizing reward without explicitly accounting for the internal costs of representing and deploying state-dependent behavior. As a result, such models can describe what an ideal learner should do, but are less suited to explaining how biological learners balance reward-seeking against the cognitive cost of flexible control.

A growing line of work has begun to address this gap by introducing information-theoretic constraints into reinforcement learning, formalizing the idea that adaptive behavior reflects a tradeoff between external reward and internal processing cost \citep{tishby2010,genewein2015bounded,grau-moya2019soft,leibfried2020mutual,LaiGershman2024_policycompression}. In this framework, \textit{policy complexity} can be quantified by the mutual information between states and actions, which measures how strongly action selection depends on state information. Low mutual information corresponds to simple, state-insensitive behavior, whereas high mutual information reflects increasingly specialized, state-contingent control. This perspective is especially appealing for cognitive science because it provides a normative language for describing bounded rationality. Agents should use state information when doing so improves reward, but only to the extent justified by the informational cost of maintaining and expressing that dependence.

In this paper, we introduce an information-constrained variant of SARSA that learns under an explicit penalty on policy complexity. Our algorithm maintains both a state-specific policy, $\pi(a\mid s)$, and an adaptive action prior $\pi(a)$, and it penalizes deviations between them through the Kullback-Leibler divergence $D_{\text{KL}}(\pi( \cdot \mid s) \| \pi(\cdot))$. Minimizing this divergence is equivalent to minimizing the mutual information between states and actions, thereby encouraging efficient policies that deviate from the prior only when the expected reward benefit is sufficient.

While information-theoretic constraints of this kind have been studied in both reinforcement learning and cognitive science, prior work has primarily used them to improve control performance and exploration \citep{grau-moya2019soft}, to derive actor-critic formulations \citep{leibfried2020mutual}, or to model human learning and generalization in one-shot decision tasks \citep{lai2021policy, fang2025humans}. Our contribution is to expand this framework to an on-policy temporal-difference (SARSA) model of sequential decision making under internal resource constraints.

Beyond its algorithmic role, this formulation provides a process-level account of how biological agents may learn reward-sensitive behavior while operating under constraints on perception, attention, and working memory. Moreover, because $D_{\mathrm{KL}}(\pi(\cdot \mid s)\|\pi(\cdot))$ quantifies the cost of selecting a state-specific action relative to a default prior, it offers a natural bridge from policy complexity to reaction time. The model therefore aims to explain not only which actions are selected, but also why choices that require greater policy differentiation may take longer, linking reward optimization, cognitive constraint, and behavioral latency within a single learning framework.

\section{Related Work}

\textbf{Info-theoretic Regularization in RL:} Information-theoretic constraints have become an established way to formalize tradeoffs between reward and computational cost in reinforcement learning. Maximum-entropy methods augment expected return with an entropy bonus \citep{ziebart2008maxent}, and related soft control formulations have led to practical algorithms such as soft Q-learning \citep{haarnoja2017softq} and soft actor-critic \citep{haarnoja2018sac}. Most directly related to the present work, \citep{grau-moya2019soft} introduce soft Q-learning with mutual-information regularization, showing that a learned marginal action prior induces a penalty on deviations between the state-conditional policy and the marginal action distribution, and \citep{leibfried2020mutual} extend this framework to mutual-information-regularized actor-critic learning. Related information-theoretic objectives have also been studied for exploration \citep{gopal2022mutual}, offline reinforcement learning \citep{ma2023mutual}, and intrinsic motivation and control \citep{aubret2023information}.

\textbf{Policy Compression and Bounded Rationality in Cognitive Science:} Closely related ideas have also been developed in cognitive science under the framework of policy compression and reward-complexity tradeoffs. \citep{lai2021policy} formalize policy complexity as the mutual information between states and actions, providing a normative account of bounded action selection and motivating a connection between policy complexity and response time. More broadly, this line of work connects to information-theoretic accounts of bounded rationality, which treat action selection as a tradeoff between expected utility and information-processing cost \citep{genewein2015bounded}. This perspective is also consistent with information-theoretic accounts of cognitive effort, in which control cost is linked to the information required to update internal models or representations \citep{zenon2019information}. Recent work shows that efficient coding can improve reinforcement-learning accounts of human generalization \citep{fang2025humans}, extending earlier work linking efficient coding to the structure of human generalization more broadly \citep{sims2018efficient}.

\textbf{Present Contribution:} Our contribution is therefore not a new information-theoretic objective, but a new use of an established one: we embed a policy-complexity penalty in a temporal-difference learning model and interpret the resulting state-wise KL-divergence cost as both a constraint on policy formation and a predictor of reaction time. This contribution also complements neuroscientific work on reinforcement learning. Actor-critic models have played a central role in biologically grounded reinforcement learning \citep{niv2009reinforcement}, but existing neural evidence does not uniquely favor a single model-free architecture. Electrophysiological studies have reported dopaminergic activity consistent with encoding action-specific value information \citep{morris2006midbrain}, and high-resolution fMRI studies have identified distinct prediction-error signals related to both state values and action values \citep{colas2017distinct}. These findings motivate broader exploration of on-policy action-value algorithms such as SARSA in computational accounts of adaptive behavior. The specific contribution of the present work is to combine policy-complexity principles with a sequential reinforcement learning process model that links reward, reaction time, and bounded choice within a single framework.

\section{Method}
\label{sec:method}

\subsection{Problem Setting and Objective}

We consider a fully observable Markov Decision Process with state space $\mathcal{S}$, action space $\mathcal{A}$, transition kernel $P(s' \mid s,a)$, reward function $r(s,a)$,  stochastic policy $\pi(a \mid s)$, and discount factor $\gamma \in [0,1)$.

Our approach augments the standard control setting with an explicit penalty on policy complexity. Following prior work on mutual-information-regularized reinforcement learning, we treat complexity as the mutual information between states and actions, which measures how strongly action selection depends on state information. The resulting control objective is
\begin{equation}
\max_{\pi} \;\left( \mathbb{E}\left[\sum_{t=0}^{T} \gamma^t r(s_t,a_t) \mid s_0 = s\right] - \beta\, I(A;S)\right),
\label{eq:mi_objective}
\end{equation}
where $\beta \geq 0$ controls the tradeoff between reward maximization and policy compression. When $\beta = 0$, the objective reduces to standard reward maximization. As $\beta$ increases, the agent is increasingly penalized for using highly state-specific action policies, encouraging simpler policies that reuse a common action prior whenever possible.

This formulation provides a compact way to model bounded sequential learning under internal resource constraints. Rather than treating the policy as an unconstrained mapping from states to actions, the objective favors policies that condition on state information only when doing so yields sufficient reward benefit. In the next subsection, we make this objective concrete by defining an adaptive marginal action prior and the corresponding state-wise complexity cost used in MI-SARSA.

\subsection{Adaptive Action Prior and Policy-Complexity Penalty}

To instantiate the policy-complexity term, we introduce a marginal action prior $p(a)$ that tracks the agent's long-run action frequencies independently of state. This prior serves as the reference distribution in the mutual-information penalty, providing a behavioral baseline against which state-conditional action preferences are compared. Intuitively, actions that are already likely under the marginal prior incur a smaller complexity cost, whereas actions that are highly specific to a particular state incur a larger one.

At initialization, $p(a)$ is uniform over $\mathcal{A}$. During learning, it is updated online using an exponential moving average of observed actions:
\begin{equation}
p_{t+1}(a) = (1-\lambda)p_t(a) + \lambda \mathbf{1}[a=a_t], \qquad \forall a \in \mathcal{A},
\label{eq:prior_update}
\end{equation}
where $\lambda \in (0,1)$ is a smoothing parameter and $\mathbf{1}[\cdot]$ is the indicator function. This update increases the mass assigned to the selected action while gradually decaying the mass assigned to all others.

Using this prior, policy complexity is measured as the mutual information between states and actions:
\begin{equation}
I(A;S) = \mathbb{E}_{s \sim d_{\pi}, a \sim \pi(a \mid s)} [\log \pi(a \mid s) - \log p(a)],
\label{eq:mutual_information}
\end{equation}
where $d_{\pi}$ is the state visitation distribution induced by $\pi$. Equivalently, this is the expected KL-divergence between the state-conditional policy $\pi(\cdot \mid s)$ and the marginal action prior $p(\cdot)$, and it therefore quantifies how much state-specific information is used to select actions.

In practice, we incorporate this complexity cost directly into the immediate reward at each time step:
\begin{equation}
\hat{r}_t = r_t - \beta \left(\log \pi(a_t \mid s_t) - \log p(a_t)\right).
\label{eq:shaped_reward}
\end{equation}
The first term encourages reward-seeking behavior, while the second penalizes deviations from the marginal action prior. Larger values of $\beta$ place greater weight on compression, encouraging policies that depart from the prior only when the expected reward gain justifies the additional informational cost. This state-wise cost term also provides the basis for our reaction-time interpretation: the same quantity that penalizes policy complexity can be read as the cost of selecting a state-specific action over a default prior.

\begin{algorithm}[tb]
\caption{MI-SARSA: Mutual Information Regularized SARSA}
\label{alg:mi-sarsa}
\begin{algorithmic}[1]
\Require Environment $\mathcal{E}$, actions $\mathcal{A}$, learning rate $\alpha$, discount $\gamma$, softmax temperature $\tau$, episodes $N$, max steps $T_\text{max}$, $\beta$, EMA rate $\lambda$
\State Initialize $Q(s,a) \gets 0 \ \forall s \in \mathcal{S}, a \in \mathcal{A}$
\State Initialize marginal action prior $p(a) \gets 1/|\mathcal{A}|$ for all $a \in \mathcal{A}$
\For{episode $= 1$ to $N$}
    \State Reset environment: $s_0 \gets \text{env.reset()}$
    \State Select initial action $a_0 \sim \pi(\cdot \mid s_0)$ using softmax over $Q(s_0,\cdot)$
    \For{$t = 0$ to $T_\text{max}-1$}
        \State Take action $a_t$, observe reward $r_t$ and next state $s_{t+1}$
        \State Update marginal prior: $p(a) \gets (1-\lambda)p(a), \quad p(a_t) \gets p(a_t) + \lambda, \quad \text{normalize } p(a)$
        \State Compute MI-regularized reward: $r'_t \gets r_t - \beta \big( \log \pi(a_t \mid s_t) - \log p(a_t) \big)$
        \State Select next action $a_{t+1} \sim \pi(\cdot \mid s_{t+1})$ using softmax over $Q(s_{t+1},\cdot)$
        \State Update Q-value: $Q(s_t,a_t) \gets Q(s_t,a_t) + \alpha \Big[ r'_t + \gamma Q(s_{t+1},a_{t+1}) - Q(s_t,a_t) \Big]$
        \State $s_t \gets s_{t+1}, \ a_t \gets a_{t+1}$
        \State \textbf{if} episode terminated \textbf{then break}
    \EndFor
\EndFor
\State \Return Learned Q-values $Q(s,a)$ and marginal prior $p(a)$
\end{algorithmic}
\end{algorithm}

\subsection{MI-SARSA}

MI-SARSA implements the objective above within a standard on-policy temporal-difference learning rule. Given a transition $(s_t,a_t,r_t,s_{t+1})$, the agent first computes the regularized reward $\hat{r}_t$ from Eq.~\eqref{eq:shaped_reward}, then samples the next action $a_{t+1} \sim \pi(\cdot \mid s_{t+1})$, and updates the action value using the standard SARSA target:
\begin{equation}
Q(s_t,a_t) \leftarrow Q(s_t,a_t) + \alpha \left[\hat{r}_t + \gamma Q(s_{t+1},a_{t+1}) - Q(s_t,a_t)\right],
\label{eq:sarsa}
\end{equation}
where $\alpha \in (0,1)$ is the learning rate. Actions are selected using a Boltzmann policy over current action values with temperature parameter $\tau > 0$. Relative to standard SARSA, the only modification is that the immediate reward is replaced by its mutual-information-regularized counterpart.

Algorithm~\ref{alg:mi-sarsa} summarizes the full procedure. At each step, the agent updates the marginal action prior, computes the state-wise complexity cost, samples the next action from the current softmax policy, and performs a SARSA update using the regularized reward. This yields a simple sequential learning rule in which action values are shaped not only by external reward, but also by the informational cost of departing from a default action prior.

\section{Results}
\label{sec:results}

\subsection{Environments}

We evaluate MI-SARSA in two tabular navigation domains that impose tradeoffs between reward, risk, and policy complexity. The first is a modified $8 \times 8$ version of Gymnasium's \citep{gymnasium} Frozen Lake environment. As in the standard task, the agent must navigate from a start state to a goal while avoiding terminal holes. We use a modified reward function where reaching the goal yields $+10$, entering a hole yields $-10$, and each step incurs a cost of $-1$.  The second is the standard $4 \times 12$ Cliff Walking domain, in which the agent must traverse from a start state to a goal while avoiding a high-penalty cliff region. Each step incurs a small cost of $-1$ and stepping into the cliff results
in a penalty of $-100$. In both tasks, the state is the agent's current grid location and the action space consists of the four primitive movements: up, down, left, and right.

These domains are useful for the present study because they contain regions that differ sharply in the consequences of action errors. States near terminal hazards or near the goal require more precise control, whereas lower-risk regions admit simpler, more reusable action policies. This structure makes both environments well suited for studying how information constraints shape policy complexity, reaction time, and adaptation under task variation.

\subsection{Evaluation Protocol}

Unless otherwise noted, we evaluate MI-SARSA by sweeping the information-regularization strength over values of $\beta$ between $0$ and $5$ and averaging results across $50$ random seeds. For the main reward-complexity analysis, each agent is trained for $1000$ episodes and then evaluated for $100$ episodes under the learned policy. We report expected return as the mean cumulative reward over evaluation episodes, and estimate policy complexity as the empirical average of $\log \pi(a \mid s) - \log p(a)$ over visited state-action pairs.

\begin{figure}[tb]
    \centering
    \begin{subfigure}[b]{0.45\textwidth}
        \centering
        \includegraphics[width=\textwidth]{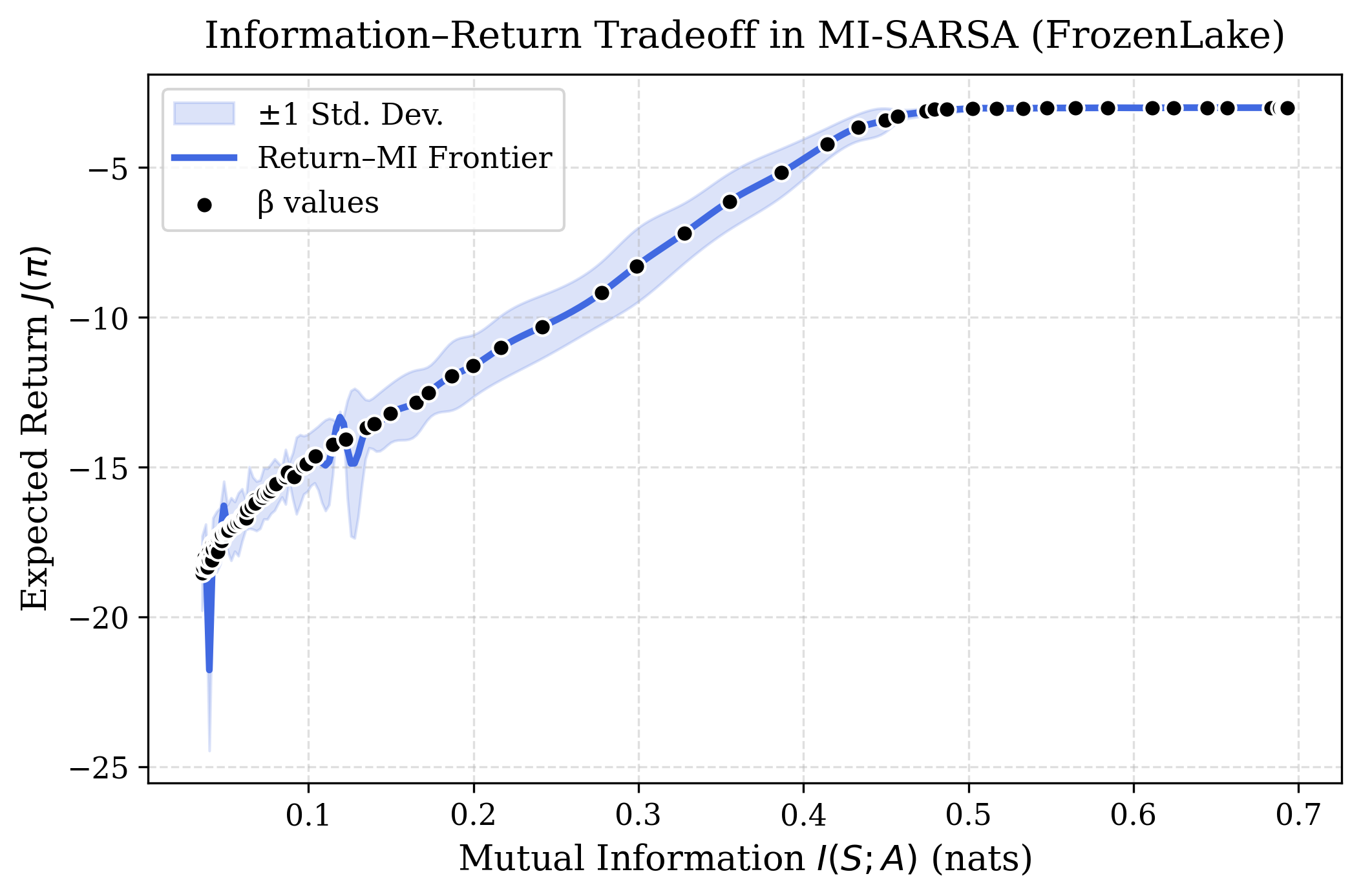}
        \caption{Frozen Lake}
        \label{fig:tradeoff_sub1}
    \end{subfigure}
    \hfill
    \begin{subfigure}[b]{0.45\textwidth}
        \centering
        \includegraphics[width=\textwidth]{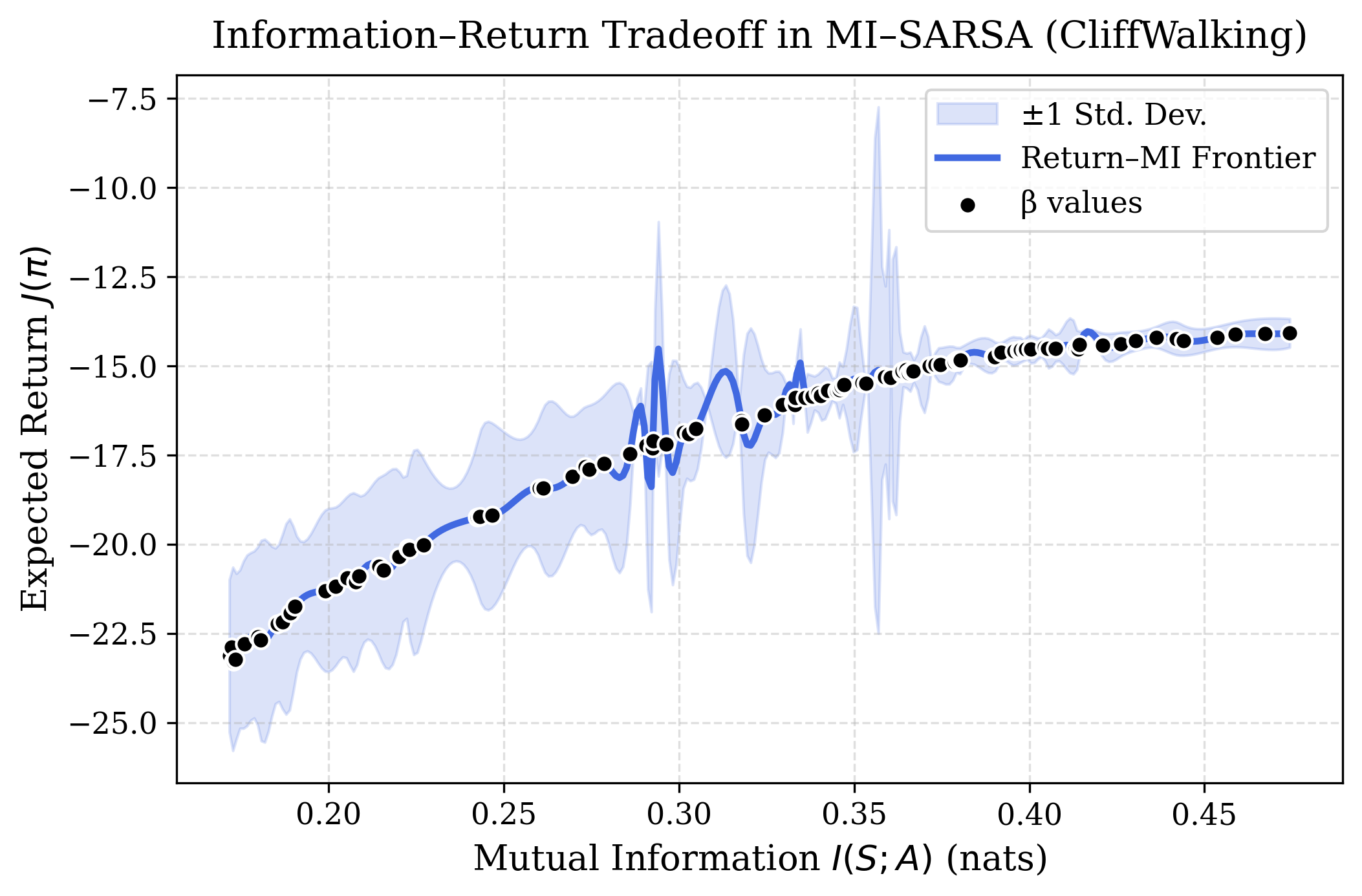}
        \caption{Cliff Walking}
        \label{fig:tradeoff_sub2}
    \end{subfigure}
    \caption{Reward-Complexity curves for Frozen Lake and Cliff Walking benchmarks using MI-SARSA. \textbf{(a):} The expected return increases as the mutual information between states and actions rises, illustrating the cost of policy complexity. Shaded regions indicate $\pm 1$ standard deviation across runs. \textbf{(b):} A similar trend is observed, with larger variance at higher complexity levels.}
    \label{fig:subfigureExample}
\end{figure}

\subsection{Reward-Complexity Tradeoff}
\label{sec:tradeoff}

Figures ~\ref{fig:tradeoff_sub1} and ~\ref{fig:tradeoff_sub2} show the empirical reward-complexity frontier induced by MI-SARSA in both domains. For each value of $\beta$, we train a policy and then evaluate its average return together with its realized policy complexity, estimated as the empirical average of $\log \pi(a \mid s) - \log p(a)$ over visited state-action pairs. Varying the regularization strength traces out a clear tradeoff between expected return and policy complexity.

Across both environments, policies with lower complexity achieve lower return, reflecting the fact that highly compressed policies rely more heavily on the marginal action prior and therefore make less use of state-specific information. As policy complexity increases, return improves, indicating that more state-dependent action selection yields better task performance. This pattern is consistent with the central prediction of mutual-information-regularized control: greater reward can be achieved, but only by paying the cost of more informative and more specialized policies.

The frontier also illustrates how information constraints shape behavioral variability. Lower-complexity policies exhibit greater variability across seeds, consistent with the use of more diffuse and less state-sensitive action distributions. Higher-complexity policies are more structured and tend to produce more stable outcomes. Taken together, these results establish that MI-SARSA recovers the expected reward-complexity tradeoff in sequential reinforcement learning, providing the basic empirical foundation for the reaction-time and environment-shift analyses that follow.

\subsection{Policy Complexity and Reaction Time}
\label{sec:rt_results}

A central prediction of the model is that the same state-specific information cost that penalizes policy complexity should also determine reaction time. Intuitively, selecting an action in a given state requires transforming the marginal action prior into a state-conditional policy. States that require little deviation from the prior can be acted on quickly, whereas states that require a highly specialized response incur a larger computational cost and should take longer. We therefore define the reaction time in state $s$ as 
$
    \text{RT}(s) = \text{RT}_0 + m D_{\text{KL}}(\pi(\cdot \mid s) \| \pi(\cdot)),
$
where $\mathrm{RT}_0$ is a baseline motor latency and $m$ sets the scale of information processing time. In the current simulations we arbitrarily set these parameters to $\mathrm{RT}_0 = 10^{-3}$ and $m = 1$.

\begin{figure}[tb]
    \centering
    \vspace{-5pt}
    
    \begin{subfigure}{\textwidth}
        \centering
        \includegraphics[width=\textwidth]{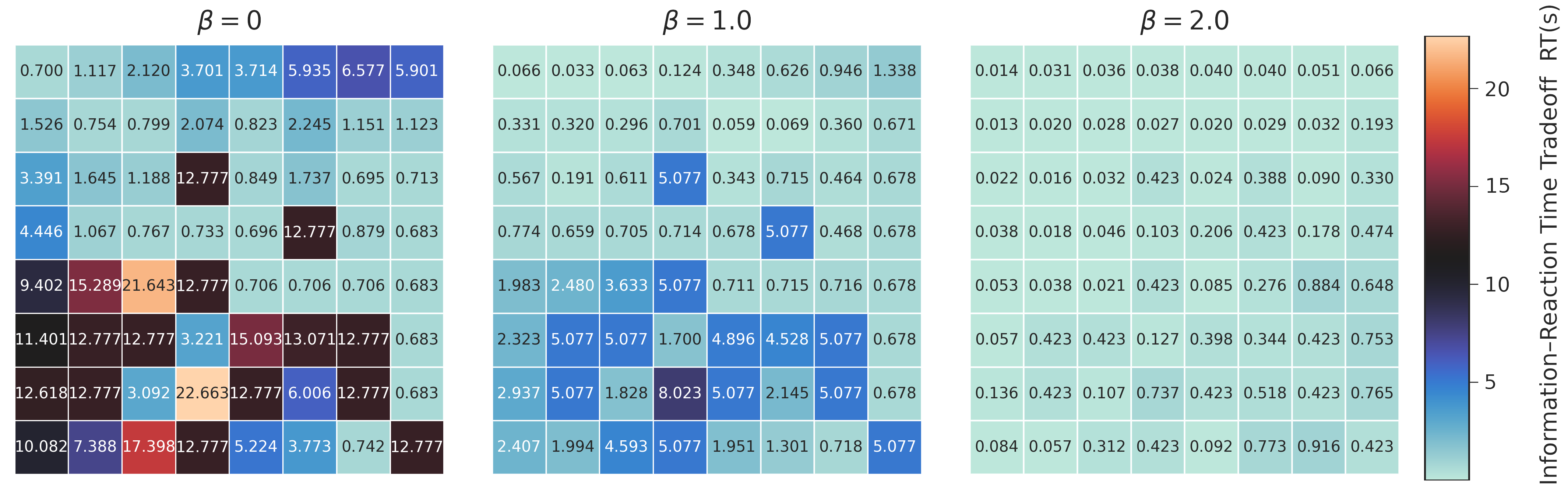}
        \caption{Modified Frozen Lake environment.}
        \label{fig:fl_heatmap}
    \end{subfigure}
    
    \vspace{4pt}
    
    \begin{subfigure}{\textwidth}
        \centering
        \includegraphics[width=\textwidth]{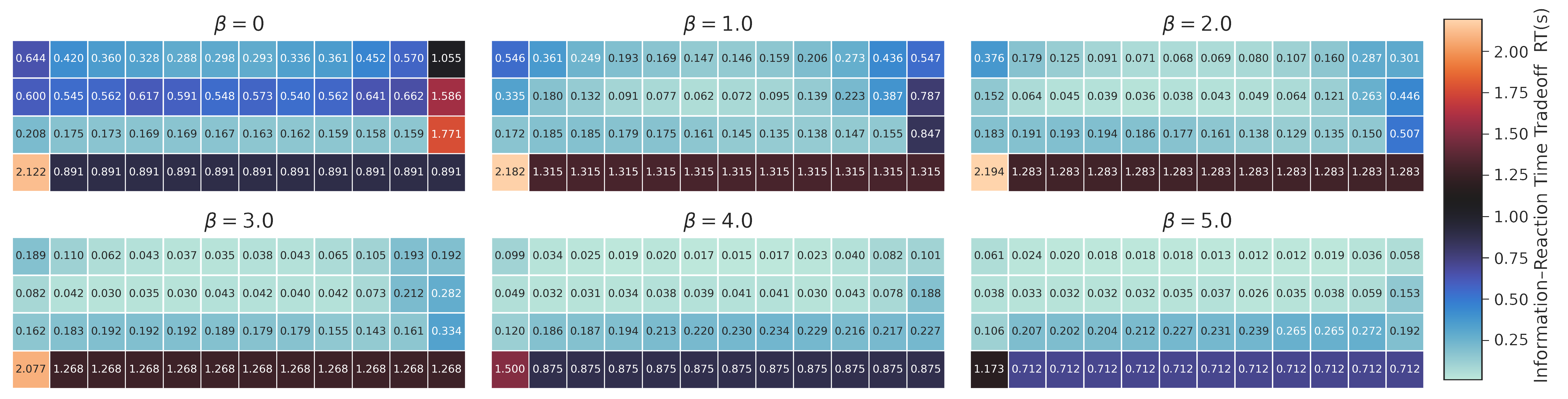}
        \caption{Cliff Walking environment.}
        \label{fig:cliff_heatmap}
    \end{subfigure}
    
    \vspace{-5pt}
    \caption{State-wise reaction time as a function of regularization strength across the two environments.}
    \label{fig:rt_heatmaps}
    \vspace{-10pt}
\end{figure}

This reaction-time interpretation is motivated by a long-standing information-theoretic view of choice latency. In Hick's classic choice reaction-time experiments \citep{hick1952rate}, mean response time increased approximately linearly with the information required to resolve the response, originally characterized by the entropy of the response set. \citep{zenon2019information} recast this idea in more general terms, arguing that cognitive cost can be formalized as the amount of information needed to update a prior belief into a context-specific posterior, quantified by Kullback-Leibler divergence. In our setting, action selection can be viewed in exactly this way: the agent begins from a marginal action prior $p(a)$ and, in state $s$, updates it to the state-conditional policy $\pi(a \mid s)$. The quantity $D_{\mathrm{KL}}(\pi(\cdot \mid s)|p(\cdot))$ therefore measures the information required to construct a state-specific action distribution from a default prior, making it a natural predictor of the time required to select an action. This extends Hick's law from a fixed response-set setting to a sequential reinforcement-learning setting in which the cost of action selection varies across states.

Figures~\ref{fig:fl_heatmap} and \ref{fig:cliff_heatmap} show that this mapping yields structured reaction-time predictions in both environments. When $\beta = 0$, reaction times are higher and more heterogeneous because the policy depends more strongly on state-specific information. As $\beta$ increases, much of the state space collapses toward baseline latency, reflecting more prior-driven behavior. However, reaction time remains elevated in behaviorally consequential states, including regions near the goal or holes in Frozen Lake and the narrow safe path above the cliff in Cliff Walking, where errors are especially costly.

Across both domains, stronger regularization reduces overall policy complexity and reaction time, while preserving slower decisions in states that require precise control. The result is a simple mechanistic link between bounded policy complexity and the spatial structure of deliberation in sequential decision making.

\subsection{Adaptation Under Environment Shift}
\label{sec:env_shift_results}

To examine how policy compression impacts performance under environment switches, we evaluate MI-SARSA on sequences of altered environments in both domains. In Frozen Lake, we generate a family of $8 \times 8$ maps consisting of one canonical map and 19 randomly sampled variants, with hole layouts drawn by sampling tile configurations with hole probability $p \sim \textrm{Uniform}(0.7, 0.9)$. In Cliff Walking, we generate one canonical map and 19 variants augmented with three monsters (terminal states) placed at randomly sampled positions, while preserving the start and goal states and ensuring that at least one column along the cliff row remains traversable. These manipulations create related tasks that preserve the overall structure of each domain while changing the local hazards the agent must navigate.

\begin{figure}[tb]
\centering
\begin{subfigure}[b]{0.245\textwidth}  
\centering
\includegraphics[width=\textwidth]{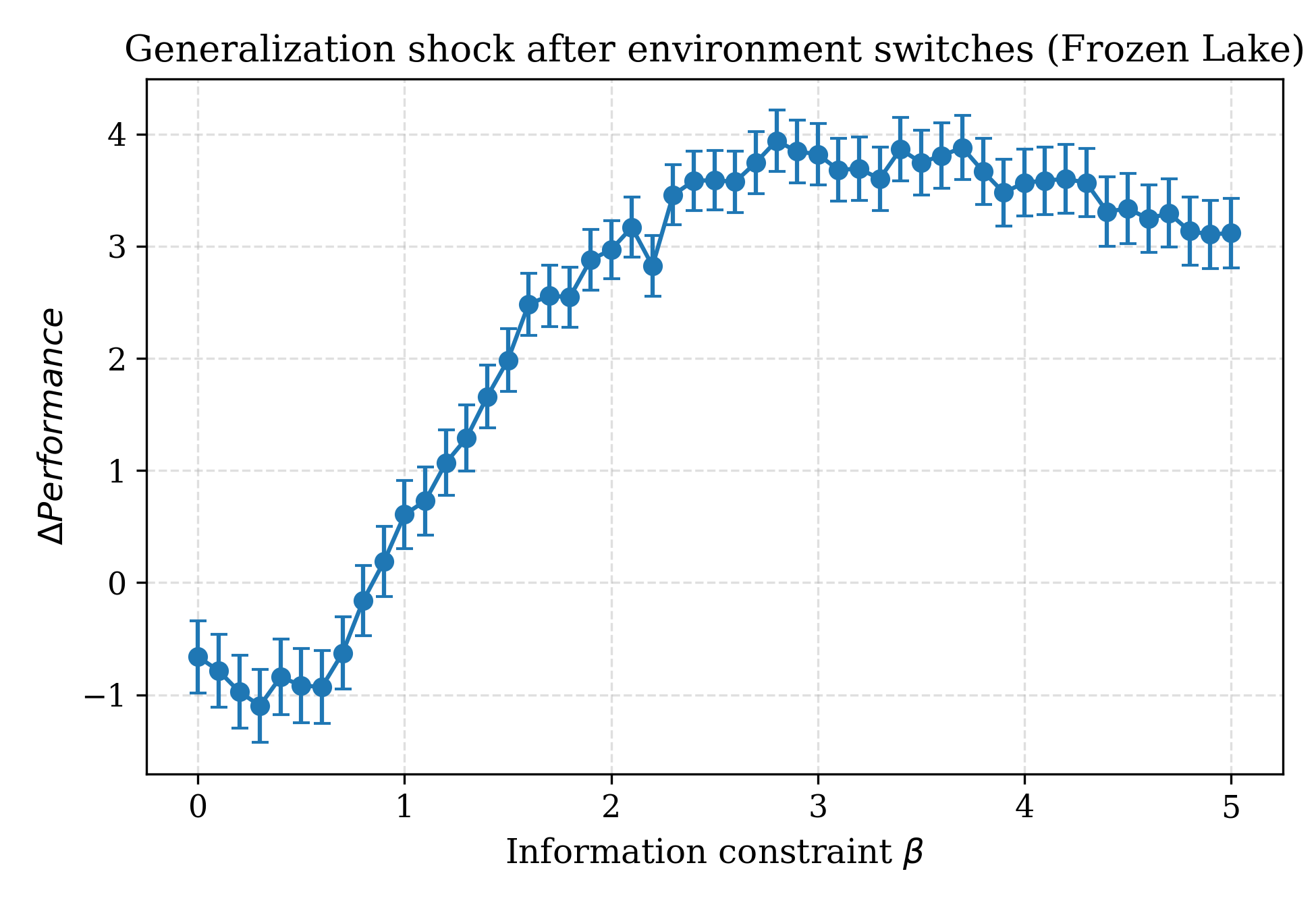}
\caption{}
\label{fig:sub1}
\end{subfigure}
\hfill
\begin{subfigure}[b]{0.245\textwidth}
\centering
\includegraphics[width=\textwidth]{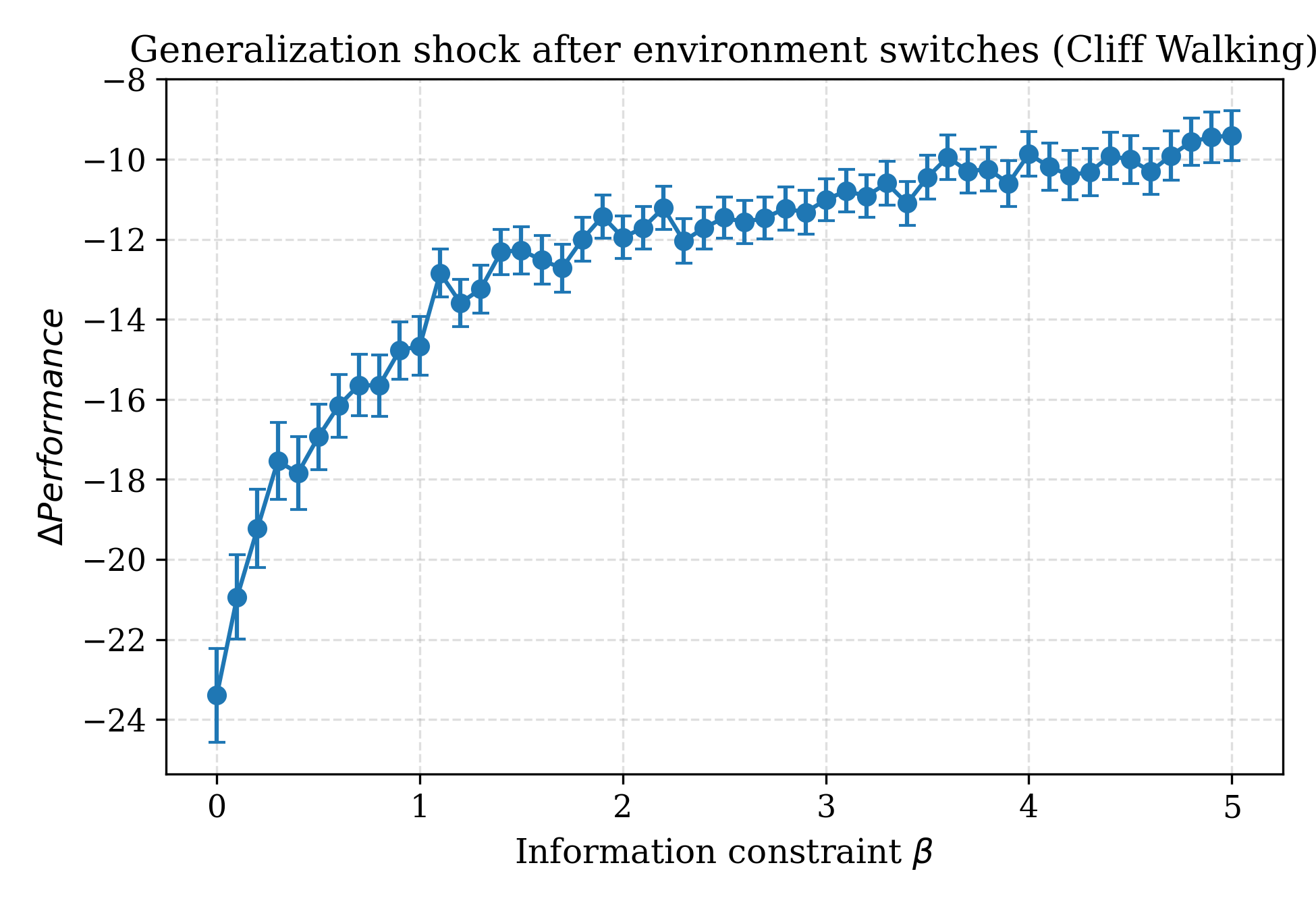}
\caption{}
\label{fig:sub2}
\end{subfigure}
\hfill
\begin{subfigure}[b]{0.245\textwidth}
\centering
\includegraphics[width=\textwidth]{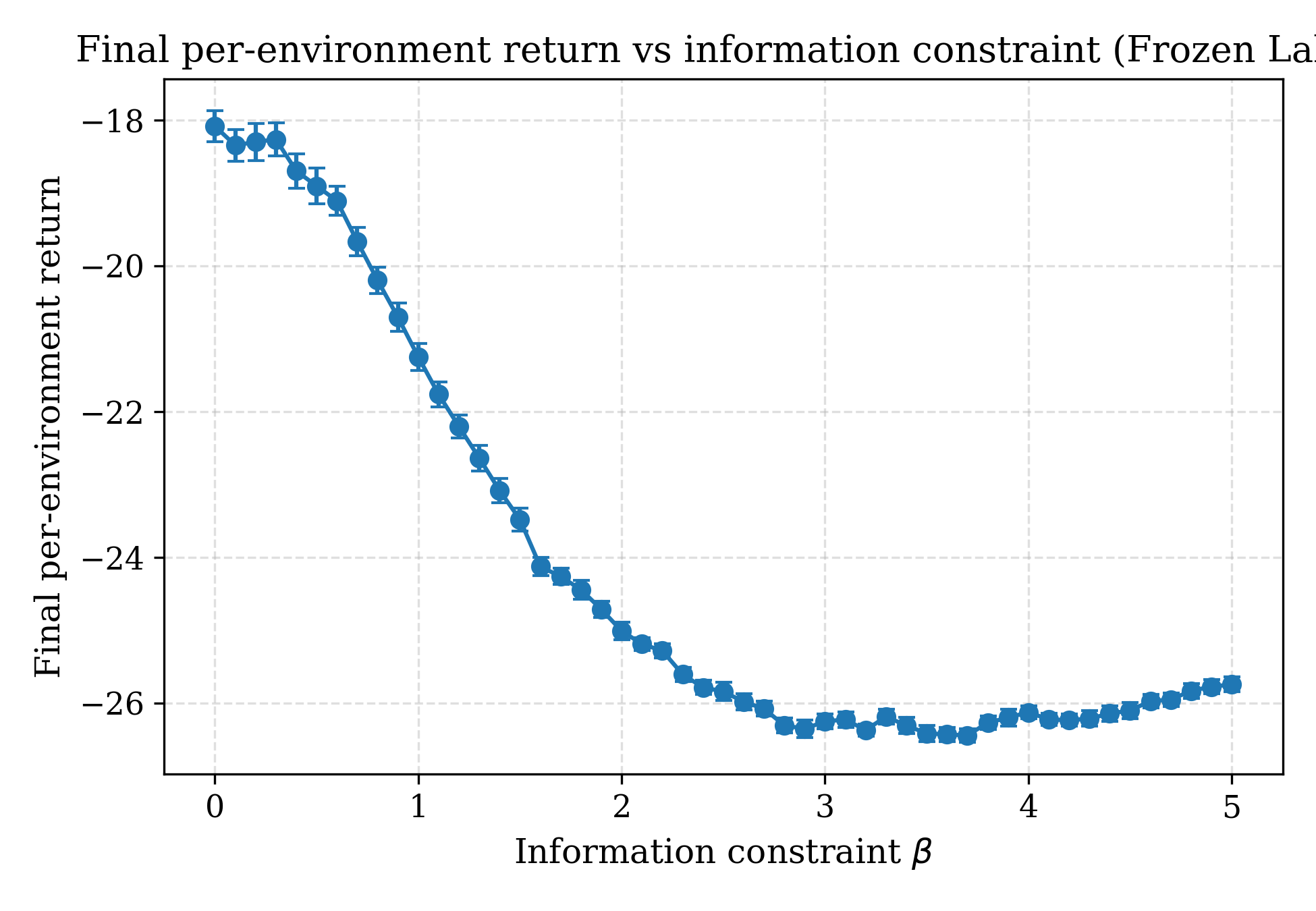}
\caption{}
\label{fig:sub3}
\end{subfigure}
\hfill
\begin{subfigure}[b]{0.245\textwidth}
\centering
\includegraphics[width=\textwidth]{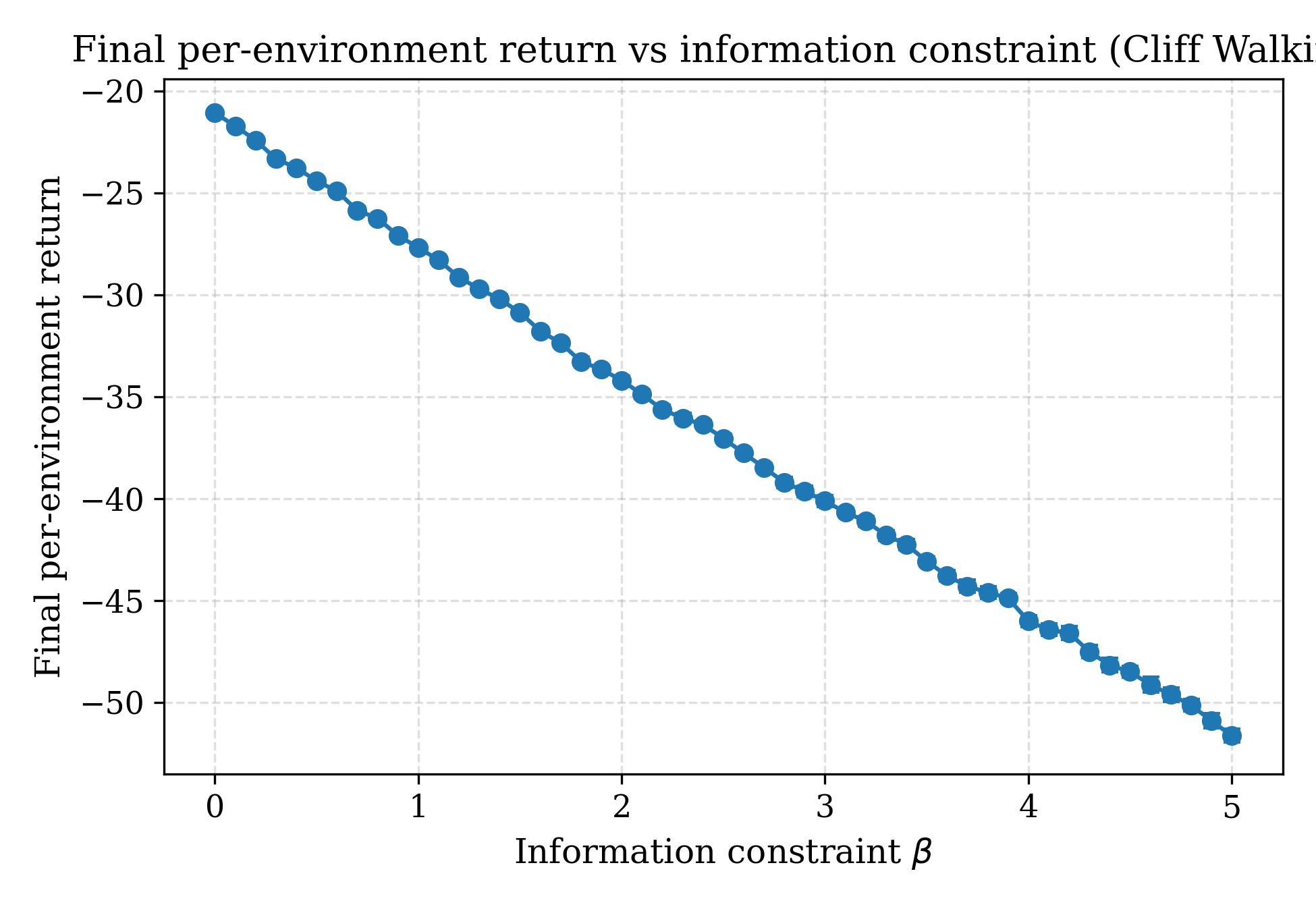}
\caption{}
\label{fig:sub4}
\end{subfigure}
\caption{
Performance degradation and post-switch return as a function of policy-complexity regularization in Frozen Lake and Cliff Walking. Panels \textbf{(a)} and \textbf{(b)} show the immediate performance drop after the switch, and panels \textbf{(c)} and \textbf{(d)} show final post-switch return, averaged over the last 20 episodes of each block. Lower regularization produces larger initial drops but higher eventual return, whereas stronger regularization reduces degradation at the cost of lower asymptotic performance.}
\label{fig:all_horizontal}
\end{figure}

For each seed, the agent is trained on a sequence of 20 environments. Training lasts 2000 episodes, divided into 20 blocks of 100 episodes, with the agent interacting with one fixed environment per block before switching to the next.  We evaluate adaptation and measure final per-environment performance as the average return over the final 20 episodes of each block over 51 values of $\beta$ between 0 and 5, averaging results across 50 seeds per value.

We quantify robustness to each environment change using $\Delta \textit{-performance}$, computed as the difference between mean performance over the initial 20 episodes of the new block and mean performance over the final 20 episodes of the preceding block. By definition, negative values indicate a deterioration in performance following the switch, whereas values near zero indicate smooth transfer, and positive values indicate improvement across the transition, suggesting policy underfitting induced by strong mutual-information constraints.

Figures~\ref{fig:sub1} and~\ref{fig:sub2} show a clear dependence of $\Delta \textit{-performance}$ on regularization strength. At low values of $\beta$, task switches produce negative values of $\Delta$, reflecting performance degradation and indicating over-specialization. In this regime, policies adapt strongly within each environment but rely heavily on task-specific structure and therefore transfer poorly to new variants. As $\beta$ increases, $\Delta \textit{-performance}$ moves toward zero, indicating smoother transitions and reduced fragility. Stronger complexity penalties encourage greater sharing of action preferences across states, reducing the extent to which the agent overfits to noisy features of a particular environment. For sufficiently large $\beta$, $\Delta$ can become positive, but this does not necessarily indicate beneficial transfer. Under strong compression, learning within each block is slower and may not saturate before the next switch, so early performance in the new block can exceed late performance in the previous one because overall learning is still improving.


This robustness, however, comes at a cost. As shown in Figure~\ref{fig:sub3} and ~\ref{fig:sub4}, final per-environment performance declines monotonically with increasing $\beta$ in both domains. Higher levels of compression reduce the expressive capacity of the policy, limiting the agent's ability to exploit detailed structure within each task. The same mechanism that dampens generalization shock therefore also constrains achievable reward.

Taken together, these results reveal a reward-robustness tradeoff governed by policy complexity. Information constraints do not produce a discrete optimal regularization level; rather, they characterize a principled tradeoff between expressivity and stability. Lower $\beta$ yields higher asymptotic performance but greater vulnerability to environment change, as evidenced by negative $\Delta \textit{-performance}$. Higher $\beta$ produces smoother environmental adaptation at the cost of reduced task-specific performance, as evidenced by less negative or even positive $\Delta$ values. This pattern supports the interpretation of policy-complexity penalties as a structural bias that mediates the balance between specialization and robustness in bounded sequential learning.

\section{Discussion}
\label{sec:discussion}

This paper develops a model of bounded sequential learning by embedding a policy-complexity cost within the SARSA temporal-difference learning rule. Across our experiments, three conclusions emerge. First, MI-SARSA exhibits the expected reward-complexity tradeoff: higher return is associated with more state-specific policies, whereas stronger regularization compresses behavior toward a simpler action prior. Second, the same state-wise Kullback-Leibler cost that governs policy compression also yields a principled account of behavioral latency, predicting where in the state space decisions should take longer. Third, moderate policy-complexity constraints can improve adaptation under environment shift by reducing over-specialization, suggesting that information constraints may act not only as limits on behavior, but also as adaptive biases that promote transfer.

These findings support an information-regularized view of bounded learning in which state information is used only when its expected benefit justifies both its computational cost and increased behavioral latency. Reward maximization, policy compression, and reaction time therefore emerge as linked consequences of the same tradeoff, providing a bridge between reinforcement-learning accounts of adaptive behavior and cognitive-science theories of limited attention, memory, and computational capacity. The present model is evaluated in simple, fully observable tabular tasks and is not fit to human data, so the contribution is best understood as a framework rather than a detailed descriptive account. Future work should test scalability to larger and partially observable problems and evaluate whether state-wise information costs predict human reaction times on a trial-by-trial basis.

\bibliographystyle{unsrt}  
\bibliography{references}  

\end{document}